# Research and Implementation of Data Enhancement Techniques for Graph Neural Networks

Jingzhao Gu
Beijing Institute of Technology
Beijing, China
gujingzhao0507@gmail.com

Haoyang Huang*
Chongqing University
Chongqing, China
* Corresponding author: hhy147scu@gmail.com

*Abstract*—Data, algorithms, and arithmetic power are the three foundational conditions for deep learning to be effective in the application domain. Data is the focus for developing deep learning algorithms. In practical engineering applications, some data are affected by the conditions under which more data cannot be obtained or the cost of obtaining data is too high, resulting in smaller data sets (generally several hundred to several thousand) and data sizes that are far smaller than the size of large data sets (tens of thousands). The above two methods are based on the original dataset to generate, in the case of insufficient data volume of the original data may not reflect all the real environment, such as the real environment of the light, silhouette and other information, if the amount of data is not enough, it is difficult to use a simple transformation or neural network generative model to generate the required data. The research in this paper firstly analyses the key points of the data enhancement technology of graph neural network, and at the same time introduces the composition foundation of graph neural network in depth, on the basis of which the data enhancement technology of graph neural network is optimized and analysed.

## I. INTRODUCTION

Deep learning models have shown remarkable success in various fields, such as image recognition, natural language processing, and speech recognition. However, one of the ma- jor challenges of deep learning models is their ability to generalize to new data. The ability of a model to generalize refers to its ability to perform well on data that it has not seen be- fore [1]. Overfitting is a common problem in deep learning models, where the model learns to fit the training data too well and fails to generalize to new data [2].

To overcome the challenge of generalization, various techniques have been proposed to improve the generalization ability of deep learning models. One of the most promising techniques is data augmentation, which involves generating new training data by applying transformations to the original data. Data augmentation has shown to be effective in im- proving the generalization ability of deep learning models and reducing overfitting [3] [4].

## II. THEORY RELATED TO GRAPH NEURAL NETWORKS

### A. image transformation

Deep learning models have achieved remarkable success in various fields such as image recognition, natural language processing and speech recognition.

recognition, natural language processing and speech recognition. However, one of the main challenges faced by deep learning models is the ability to generalise to new data [5]. The generalisation ability of a model's model refers to its ability to perform well on unseen data. Overfitting is a common problem in deep learning models, i.e. overfitting of the model during learning is a common problem in deep learning models, i.e. overfitting of the model to the training data during the learning process, which prevents it from generalising to new data. Convolutional neural networks are very widely used in image recognition, due to the design characteristics of the network, which makes the network unable to achieve spatial transformation invariance to the input image. The convolutional neural network correctly recognised the original image, but when the original image was rotated by a certain angle, the recognition result of the convolutional neural network showed some unexplainable errors [6] [7][8].

### B. Basic Concepts of Graph Neural Networks

Graph structures are ubiquitous in life and are a typical data structure. A graph is represented as $G = \{V, E\}$, where V is the set of nodes and E is the set of edges in the graph structure. $v_i \in V$ denotes a node in the graph structure, and $e_{ij} = (v_i, v_j)$ denotes an edge connecting nodes $V_i$ and $V_j$. $N(v_i) = v_j | (v_i, v_j) \in E$ denotes the set of neighbours of node $V_i$.

### C. Common graph neural networks

Graph neural networks have generated many variants based on different ways of message construction and information transfer, and several common forms of variants are described next.

(1) Graph Convolutional Network (GCN)

A graph convolutional network is similar to a convolutional neural network and is actually a feature extractor for graph structures. The graph convolutional network is a first-order local approximation of the spectral graph convolution, each convolutional layer only deals with the first-order neighbourhood information, and the information transfer to the multi-order neighbourhood is achieved by stacking several convolutional layers. The graph convolutional network structure is shown in Figure 1.

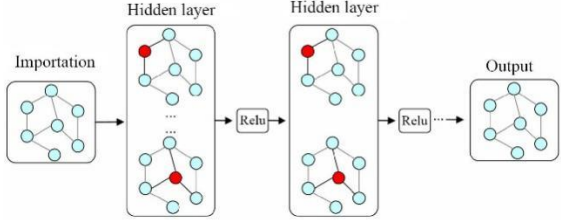

Figure 1.　Graph Convolutional Network Structure

The inputs to the graph convolutional network are the node information vector matrix H and the adjacency matrix A. The computation of each convolution is shown in Equation 1.

$$H^{(l+1)} = \sigma(\tilde{D}^{-1/2}\tilde{A}\tilde{D}^{-1/2}H^{(l)}W^{(l)}) \quad (1)$$

where $\sigma$ is the nonlinear activation function; $W^{(l)}$ is the linear transformation matrix of layer l; $\tilde{A}$ is the adjacency matrix with added self-connections; and $\tilde{D}$ the degree matrix of the adjacency matrix.

(2) Graph Saple and Aggregation Network (GraphSAGE)

In order to solve the problem that graph convolutional networks cannot handle inductive tasks, graph sampling and aggregation networks were proposed. The graph sampling and aggregation network structure is shown in Fig. 2.

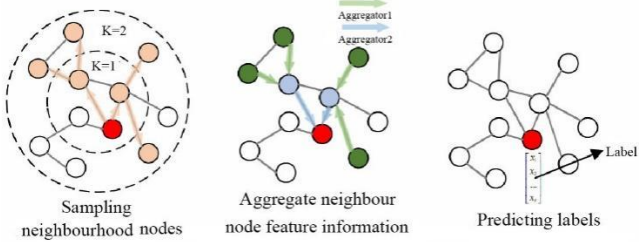

Figure 2.　Graph sampling and aggregation network structure

The graph sampling and aggregation network is divided into three parts, firstly neighbouring nodes are sampled, then neighbouring nodes are continuously fused through a multilayer aggregation function, and finally the fused node information is used to learn against the downstream task. In the sampling phase, a number k of sampling neighbour nodes is defined, and when the number of neighbour nodes of the target node is greater than k, k neighbour nodes are obtained through the sampling strategy to be included in the set of neighbour nodes of the target node. If the number of neighbouring nodes of the target node is less than k, there is put back to resample until k nodes are sampled. The specific calculation is shown in Equation 2.

$$h_v^{(l)} = \sigma(W^{(l)} \cdot CONCAT(h_v^{(l-1)}, Aggregation(\{m_u^{(l)}, u \in N(v)\}))) \quad (2)$$

Where $h_v^{(l)}$ is the representation of node v in layer l and $\sigma$ is the nonlinear activation function. Firstly all the neighbouring node message information of node v is calculated $m_u^{(l)}$, and then aggregation of messages is performed. In order to avoid node v's own information being lost during the transmission process, the output $h_v^{(l-1)}$ of node v's previous layer is spliced with the result of neighbour aggregation through the CONCAT operation, and then the node's output vector of the current layer is obtained through a linear mapping and activation function. In contrast to graph convolutional networks, graph sampling and aggregation networks introduce three message aggregation functions, namely average aggregation, maximum pooling aggregation and LSTM aggregation.

(3) Graph Attention Network (GAT)

Each node is assumed to be equally important in graph convolutional networks, thus ignoring the importance of relationships between nodes and the importance of each neighbouring node to the target node. The attention mechanism has been successfully applied to image processing, machine translation and other fields. Researchers have combined the attention mechanism with graph convolutional networks to propose graph attention networks [9]. The structure of the graph attention layer is shown in Figure 3.

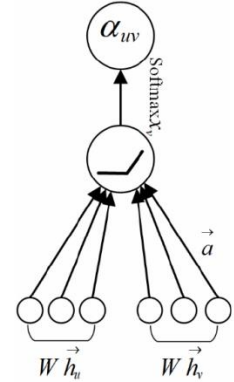

Figure 3.　Attention layer structure

The graph attention network achieves the update of node representation by stacking graph attention layers. The specific calculation process is shown in Equations 3 and 4.

$$\alpha_w = \frac{exp(Att(h_w^{(l-1)}, h_w^{(l-1)}))}{\sum_{k \in N(v)} exp(Att(h_w^{(l-1)}, h_k^{(l-1)}))} \quad (3)$$

$$h_v^{(l)} = \delta(\sum_{u \in N(v)} \alpha_{uv} W^{(l)} h_u^{(l-1)}) \quad (4)$$

Where, $\alpha_{uv}$ denotes the attention coefficient from node u to node v, $\delta$ is the activation function, and N(v) denotes the set of neighbouring nodes of node v. Att($\cdot$) is the method of attention computation, and the common way is $Leakyrelu(a^T[W^{(1)}h_v^{(l-1)} \| W^{(l)}h_u^{(l-1)}])$, where, $W^{(l)}$ is the node vector line transformation weight matrix, $a^T$ is the learnable parameter, and || denotes the splicing operation of node vectors. The representation $h_v^{(l)}$ of the final updated node v in layer l can be obtained by Eq. 4.

*D. Graph Similarity Learning*

There are two main learning ideas for graph similarity learning: one is graph embedding, where a vector representation of each graph is obtained and the similarity between different graphs is measured by the distance of the graph vectors. The advantage of this approach is that if there is a need to compare a large number of graph structures in a database, it is sufficient to compare the graph vectors in the database using some of the existing distance calculation methods. The other is to perform pairwise comparison of graphs, focusing on the similarity of individual nodes between two graphs, which is more accurate but comes with high computational complexity. The flow of the two approaches is shown in Figure 4.

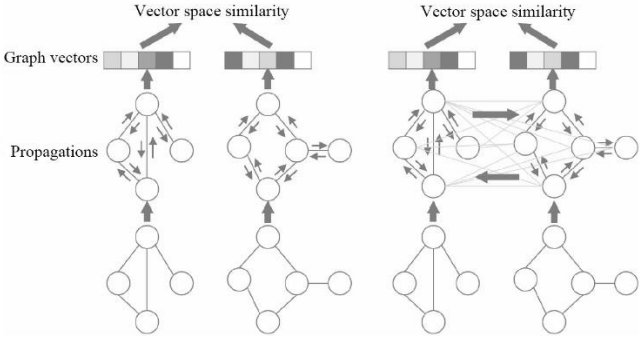

Figure 4. Figure Similarity Calculation Flow

Specifically, given first two graph structures $G_1 = (V_1, E_1)$ and $G_2 = (V_1, E_1)$ each of which V and E represent sets of nodes and sets of edges, each node v has a node feature vector $h_v$. The task of graph similarity learning is to design a model f to generate a similarity score for two graphs, i.e. $h_v$. The task of graph similarity learning is to design a model f to generate a similarity score for two graphs, i.e. $score = f(G_1, G_2)$.

III. GRAPH DATA AUGMENTATION EXPERIMENTAL RESULTS

In this section, we present the experimental results that demonstrate the effectiveness and improvement achieved by our proposed Graph Data Augmentation (GDA) methods: FDM and FANA [10]

*A. Supervised Node Classification with GDA*

Table 1 FDM and FANA performance across 3 GNN architectures and 6 benchmark datasets. Red numbers represent the top 2 best performances

| Model | Method | CORA | CITESEER | PPI | BLOGCATALOG | FLICKR |
|---|---|---|---|---|---|---|
| GCN | I | 81.05±1.45 | 71.10±1.00 | 56.00±2.00 | 76.85±0.85 | 51.62±0.36 |
| | ER | 81.35±1.35 | 71.75±1.35 | 67.95±1.45 | 78.10±2.70 | 51.25±0.42 |
| | FM | 82.10±1.30 | 72.75±1.65 | 55.75±4.05 | 76.75±2.75 | 51.69±0.49 |
| | ND | 80.90±1.30 | 71.20±1.10 | 91.60±2.10 | 93.50±1.40 | 47.79±0.35 |
| | RWS | 81.50±0.80 | 70.90±1.40 | 85.55±0.75 | 90.35±0.85 | 49.13±0.48 |
| | FDM | 82.6±0.60 | 71.1±0.85 | 58.9±1.05 | 65.05±0.55 | 47.79±0.25 |
| | FANA | 83.0±0.90 | 72.0±0.95 | 56.3±1.40 | 78.25±0.55 | 51.80±0.32 |
| GSAGE | I | 80.65±0.85 | 70.10±1.00 | 57.65±1.65 | 78.35±2.85 | 51.82±1.27 |
| | ER | 83.10±1.00 | 72.75±0.85 | 62.40±1.30 | 78.60±2.80 | 51.81±1.02 |
| | FM | 83.80±1.10 | 73.55±0.85 | 58.60±1.70 | 77.85±2.65 | 52.03±1.20 |
| | ND | 82.75±0.85 | 72.90±1.30 | 85.35±0.85 | 88.50±0.50 | 50.35±0.96 |
| | RWS | 79.30±1.00 | 68.05±0.85 | 78.15±1.15 | 85.90±0.90 | 51.28±1.30 |
| | FDM | 80.8±0.55 | 69.6±0.40 | 61.6±0.70 | 76.80±0.90 | 42.26±0.93 |
| | FANA | 81.2±0.65 | 70.0±0.40 | 57.85±1.55 | 77.1±0.70 | 52.78±0.51 |
| GAT | I | 70.75±3.75 | 60.10±2.70 | 82.50±1.80 | 92.30±0.70 | 52.55±0.41 |
| | ER | 69.70±2.80 | 59.55±2.85 | 83.90±1.20 | 92.00±1.00 | 52.14±0.39 |
| | FM | 71.10±2.80 | 60.05±3.35 | 82.55±1.25 | 92.40±0.90 | 52.35±0.53 |
| | ND | 69.05±2.95 | 57.65±3.15 | 83.00±1.10 | 91.90±0.60 | 52.00±0.80 |
| | RWS | 70.15±2.55 | 56.65±2.15 | 82.85±1.55 | 91.85±0.75 | 51.31±0.55 |
| | FDM | 75.3±0.80 | 62.6±0.95 | 83.0±0.25 | 92.50±0.30 | 48.9±0.09 |
| | FANA | 76.5±1.10 | 64.4±0.80 | 80.6±0.80 | 90.00±0.70 | 52.1±0.32 |

Note:
| | |
|---|---|
| I: | Identity, the original graph without data augmentation |
| ER: | Edge Removing [11]. |
| FM | Feature Masking [11]. |
| ND | Node Dropping [11]. |
| RWS | Random Walk Sampling [11]. |
| FDM | Feature Augmentation with Degree Multiplication |
| FANA | Feature Aggregation with Normalized Adjacency |

In this section, we present the results of the experiment, which is shown in Table 1, and analyze the findings. We compare the accuracy of the GNN models trained with and without data augmentation. We discuss the implications of the results and analyze the effectiveness of the proposed data augmentation.

We first evaluate our two proposed methods for node classification using supervised learning on the benchmark TUDataset. In addition, the accuracy of training 3 GNN mod- els with augmentations across datasets is reported in Table 1. The red bolded part is the best two experimental results under this dataset. The blue bolded part shows the improve- ment of our proposed two data augmentation algorithms compared to the experiment where no data augmentation is applied.

Overall, the experimental results demonstrate the superiority of FDM and FANA in terms of their performance across different GNN architectures, and datasets, and when compared to alternative data augmentation methods. These findings emphasize the effectiveness of the proposed approaches in improving the accuracy of supervised node classification tasks.

## B. Self-Supervised Graph classification with GDA via Contrastive Learning

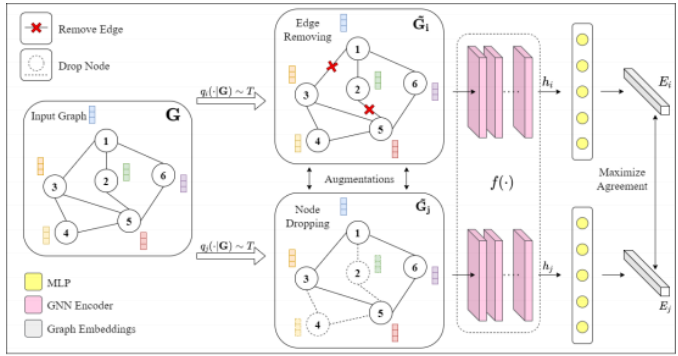

Figure 5. The framework of our finetuned graph contrastive learning. Two graph augmentations $q_i(\cdot|G)$ and $q_j(\cdot|G)$ are sampled from an augmentation pool $T$ and applied to input graph $G$. A shared GNN-based encoder $f(\cdot)$ and a MLP are trained to maximize the agreement between representations $z_i$ and $z_j$ via a contrastive loss.

In Figure 6, pairing "Identical" stands for a no-augmentation baseline for contrastive learning. Warmer colors indicate better performance gains. The baseline F1-score are 71.92%, 72.41%, 71.25%, and 85.28% for the four datasets respectively.

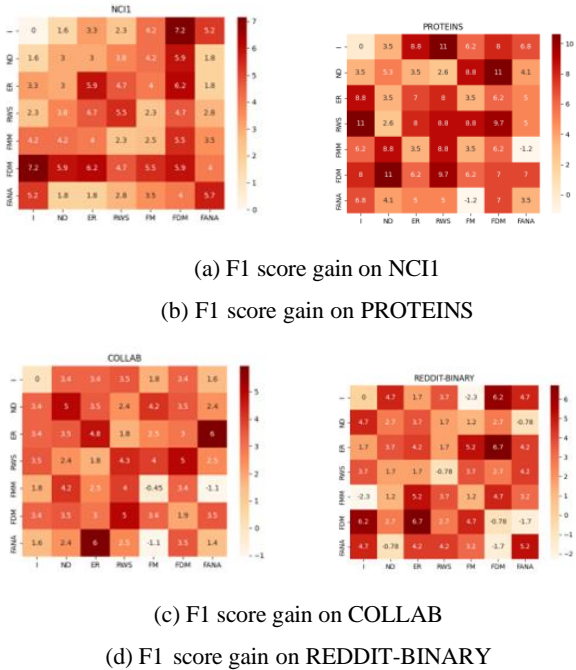

(a) F1 score gain on NCI1

(b) F1 score gain on PROTEINS

(c) F1 score gain on COLLAB

(d) F1 score gain on REDDIT-BINARY

Figure 6. Self-supervised learning F1 score gain (%) when contrasting different augmentation pairs us- ingGIN, compared to pairs that are all original graphs under four datasets: NCI1, PROTEINS, COLLAB, and REDDIT-B.

## IV. CONCLUSIONS

We explored various aspects of data augmentation (DA) in the field, focusing on both Euclidean and non-Euclidean data. Techniques and algorithms for augmenting image data, including basic manipulations, Autoaugment, and GAN-based approaches, are investigated. Additionally, novel algorithms, namely FDM and FANA, are proposed and evaluated for graph data augmentation (GDA).

In conclusion, this thesis presents a comprehensive analysis of data augmentation tech- niques for both Euclidean and non-Euclidean data. The findings underscore the importance of tailored approaches in improving the performance and generalization of deep learning models. The proposed FDM and FANA algorithms demonstrate superior performance in graph data augmentation tasks, showcasing their potential for further advancements in this area. Overall, this research contributes to the broader understanding of data augmentation and paves the way for future research and innovation in the field.

## V. APPENDIX

### A. Proposed FDM Algorithm

In this section, we introduce the FDM, which is a novel data augmentation framework for GNNs that expands the node features of nodes with higher degrees appropriately.

**Motivation** The motivation behind FMD is to handle the problem of super nodes in GNNs. In GNNs, neighborhood propagation is the core idea, but after one neighborhood propagation, the super nodes can cause feature pollution to all the neighboring nodes. This problem can lead to devastating results as the features of neighboring nodes become indistinguishable. To solve this problem, FMD uses a smooth function curve pair to handle the multipliers, where the Sigmoid function is found to be the most suitable function curve with better results as the Sigmoid function has a curve that gradually decreases the derivative of independent variable x as x increases, which is suitable for the current situation.

**Theorem** Let $G = (V, E)$ be an undirected graph with node feature matrix $X \in R^{N \times F}$, where $N$ is the number of nodes and $F$ is the number of features. Let $D \in R^{N \times N}$ be the diagonal degree matrix and α be the parameter that controls the range of $X$. Then, the augmented feature matrix $X_{aug} \in R^{N \times F}$ obtained by applying FDM is given by Eq. 5.

$$X_{aug} = X \cdot D \cdot \frac{1}{1+e^{-\alpha D}} \quad (5)$$

**FDM Algorithm Description** The algorithm takes as input a graph $G$, represented by its set of nodes $V$, and edges $E$. Additionally, it requires a node feature matrix $X$, which contains the features of each node, where $N$ is the number of nodes and $F$ is the number of features per node. The degree matrix $D$, is also provided, which represents the degree of each node in the graph. Lastly, the algorithm requires a parameter α, that controls the degree-based multiplier.

**Algorithm 1** FDM Algorithm
1: **Input:** Graph $G = (V, E)$, node feature matrix $\mathbf{X} \in \mathbb{R}^{N \times F}$, degree matrix $\mathbf{D} \in \mathbb{R}^{N \times N}$, parameter $\alpha$
2: **Initialize:** Augmented feature matrix $\mathbf{X_{aug}} \in \mathbb{R}^{N \times F}$
3: **Compute:**
4: **for** each node $v_i$ in $V$ **do**
5:     Calculate the degree $d_i$ of node $v_i$ using $\mathbf{D}$
6: **end for**
7: **for** each node $v_i$ in $V$ **do**
8:     Compute the multiplier $m_i$ for node $v_i$ using the Sigmoid function:
$$m_i = \frac{1}{1+e^{-\alpha d_i}}$$
9: **end for**
10: **for** each node feature $x_{ij}$ in $\mathbf{X}$ **do**
11:     Augment the feature using the multiplier:
$$x'_{ij} = x_{ij} \cdot d_i \cdot m_i$$
12: **end for**
13: **Output:** Augmented feature matrix $\mathbf{X_{aug}}$

Shown in above Pseudocode, the algorithm starts by initializing an augmented feature matrix, $X_{aug}$, which will contain the augmented features for each node.

Next, the algorithm computes the degree of each node in the graph by iterating over each node $v_i$ in $V$. This is done using the degree matrix $D$.

Then, for each node $v_i$ in $V$, the algorithm computes a multiplier $m_i$, using the Sigmoid function, presented in Eq. 6.

$$m_i = \frac{1}{1+e^{-\alpha d_i}} \quad (6)$$

Here, $d_i$ represents the degree of node $v_i$ calculated in the previous step.

After obtaining the multipliers for all nodes, the algorithm proceeds to augment the features of each node in the node feature matrix $X$. For each feature $x_{ij}$ of each node, the augmented feature $x'_{ij}$, is computed as Eq. 7.

$$x'_{ij} = x_{ij} \cdot d_i \cdot m_i \quad (7)$$

This equation applies the multiplier $m_i$, along with the degree $d_i$, to the original feature $x_{ij}$.

Finally, the algorithm outputs the augmented feature matrix $X_{aug}$, which contains the augmented features for each node.

In summary, the FDM algorithm enhances the node features of a graph by augmenting them using degree-based multipliers. The degree of each node is calculated, and a multiplier is computed for each node using the Sigmoid function with a user-defined parameter, α. The node features are then augmented by multiplying them with the degree and multiplier, resulting in an augmented feature matrix.

### B. Proposed FANA Algorithm

In this section, we introduce the FANA framework which is enlightened by the well-known graph convolutional network (GCN). FANA stands for Feature Aggregation with Normalized Adjacency whose core idea is to directly and fully utilize the adjacency information between nodes in the data augmentation phase in order to achieve augmentation of node features. By leveraging this label-free information, we can consistently realize improvements in test performance in supervised node classification tasks and self-supervised graph classification tasks across augmentation settings, GNN architectures, and datasets.

**Motivation** Data augmentation has been proposed as a way to improve the performance of graph representation learning models, by creating additional training data that can help the model learn to be more robust to variations in the input data. However, existing data augmentation techniques such as feature shuffling and feature masking have limitations, as they do not fully utilize the adjacency information between nodes in the graph.

In response to this challenge, we propose FANA, a novel data augmentation technique that directly and fully utilizes the adjacency information between nodes in the graph to augment node features. FANA works by computing a normalized adjacency matrix for the graph, which represents the relationships between nodes in a way that is independent of the specific feature values. FANA then aggregates the feature values with a probability for each node's neighbors in the graph, weighted by the corresponding entries in the normalized adjacency matrix, to create a new set of augmented feature values.

By directly incorporating the adjacency information in the data augmentation phase, FANA can help the model learn to better capture the relationships between nodes in the graph, leading to improved performance on a range of graph learning tasks. Furthermore, FANA is computationally efficient and can be easily integrated into existing graph representation learning pipelines.

**Theorem** Let $G = (V, E)$ be an undirected graph with node feature matrix $X \in R^{N \times F}$, where $N$ is the number of nodes and $F$ is the number of features. Let $A \in R^{N \times N}$ be the adjacency matrix and $D \in R^{N \times N}$ be the diagonal degree matrix and $p \in [0,1]$ be the probability of applying feature aggregation. Then, the augmented feature matrix $X_{aug} \in R^{N \times F}$ obtained by applying FANA is given by Eq. 8.

$$X_{aug} = p \cdot D^{-\frac{1}{2}} A D^{-\frac{1}{2}} \cdot X + (1-p) \cdot X \quad (8)$$

**Algorithm 2** FANA Algorithm
1: **Input:** Node feature matrix $\mathbf{X} \in \mathbb{R}^{N \times F}$, adjacency table $\mathbf{E}$
2: Add a self-loop for each node in the graph
3: Compute adjacency matrix $\mathbf{A} \in \mathbb{R}^{N \times N}$ from adjacency table $\mathbf{E}$
4: Compute degree matrix $\mathbf{D} \in \mathbb{R}^{N \times N}$ from adjacency matrix $\mathbf{A}$
5: Compute normalized adjacency matrix $\widetilde{\mathbf{A}} \leftarrow \mathbf{D}^{-\frac{1}{2}} \mathbf{A} \mathbf{D}^{-\frac{1}{2}}$
6: **for** $i = 1$ to $N$ **do**
7: Generate random number $r \in [0, 1]$
8: **if** $r \leq p$ **then**
9: Compute aggregated feature vector $x_{i,aug} \leftarrow \widetilde{\mathbf{A}} \mathbf{X}_{i,:}$
10: Update augmented feature matrix $\mathbf{X}_{aug,i,:} \leftarrow x_{i,aug}$
11: **else**
12: Update augmented feature matrix $\mathbf{X}_{aug,i,:} \leftarrow \mathbf{X}_{i,:}$
13: **end if**
14: **end for**
15: **Output:** Augmented feature matrix $\mathbf{X}_{aug} \in \mathbb{R}^{N \times F}$

Note: $X_{i,:}$ denotes the feature vector of node $i$ in the node feature matrix $X$, and $X_{aug,i,:}$ denotes the corresponding augmented feature vector in the augmented feature matrix $X_{aug}$. The operation $\tilde{A} X_{i,:}$ computes the aggregated feature vector for node $i$ using the normalized adjacency matrix $\tilde{A}$.

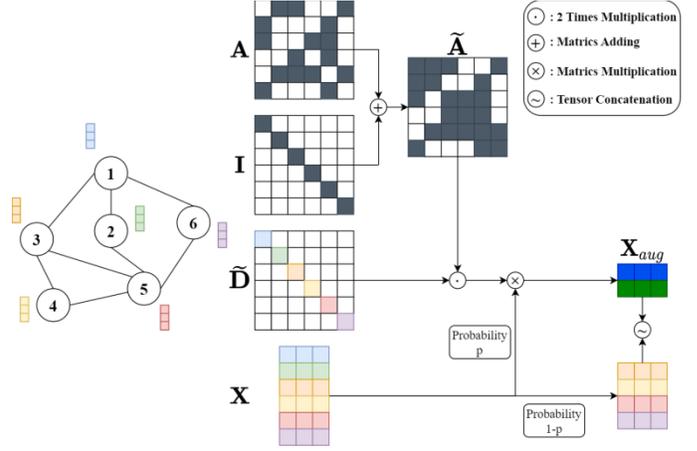

Figure 7. FANA Framework. Firstly, adjacency matrix $A \in R^{N \times N}$, diagonal degree matrix $D \in R^{N \times N}$ as well as the node feature matrix $X \in R^{N \times F}$ are obtained from the graph. Then we add a self-loop for all the nodes by adding Unit Matrix $I \in R^{N \times N}$ to $A$ and do symmetric normalization on $A$ to get $\tilde{A}$. The input feature matrix $X$ will be aggregated by $\tilde{A}$ with a probability $p$ to get the final output $X_{aug} \in R^{N \times F}$.

## C. Dataset description

**Graph dataset for supervised learning** The datasets include CORA, CITESEER, PPI, BLOGCATALOG, and FLICKER. Each dataset contains nodes representing different entities and edges representing relationships between them. Each node has a set of features, and the task is to predict the category of the node based on its features and the relationships with other nodes in the graph. Table 2 is the summary statistics and experimental setup for the six datasets.

Table 2 Summary statistics for the 5 datasets used in Supervised Node Classification with GDA

| Content | Dataset | | | | |
|---|---|---|---|---|---|
| | CORA | CITESEER | PPI | BLOGCATALOG | FLICKER |
| Nodes | 2078 | 3327 | 10076 | 5196 | 575 |
| Edges | 5278 | 4522 | 157213 | 171743 | 239738 |
| Features | 1433 | 3703 | 50 | 8189 | 12047 |
| Classes | 7 | 6 | 121 | 6 | 9 |
| Category | Literature references | | Bioinformatics | Social network | |

**Graph dataset for Self-Supervised Graph Classification** we conducted evaluations using eight diverse datasets to assess the performance of our proposed graph data augmentation methods, shown in Table 3.

Table 3 Summary statistics for the 8 datasets used in Self-Supervised Graph classification with Graph Data Augmentation via contrastive learning

| Content | Dataset | | | | | | | |
|---|---|---|---|---|---|---|---|---|
| | NCI1 | MUTAG | PROTEINS | DD | COLLAB | REDDIT-B | REDDIT-MULTI-5K | IMDB-B |
| Graphs | 4110 | 188 | 1113 | 1178 | 5000 | 2000 | 4999 | 1000 |
| Classes | 2 | 2 | 2 | 2 | 3 | 2 | 5 | 2 |

| Avg. Nodes | 29.87 | 17.93 | 39.06 | 284.32 | 74.49 | 429.63 | 508.52 | 19.77 |
| --- | --- | --- | --- | --- | --- | --- | --- | --- |
| Avg. Edges | 32.30 | 19.79 | 72.82 | 715.66 | 2457.78 | 497.75 | 594.87 | 96.53 |
| Category | Small molecules | | Bioinformatics | | | Social networks | | |